\documentclass[sigconf,table,xcdraw,nonacm]{acmart}
\usepackage[utf8]{inputenc}
\usepackage{multirow}   
\usepackage{colortbl}   
\usepackage{latexsym}
\usepackage{graphicx}
\usepackage{hyperref}
\usepackage{url}
\usepackage{enumitem}
\usepackage{multirow}
\usepackage{booktabs}
\usepackage{subcaption}
\usepackage{pifont}
\usepackage{amsmath}
\usepackage{booktabs}
\usepackage{tabularx}
\usepackage[T1]{fontenc}
\usepackage{mathtools}
\usepackage[utf8]{inputenc}
\usepackage{algorithm}
\usepackage{algorithmic}
\usepackage{microtype}

\usepackage{graphicx}

\definecolor{highlightblue}{rgb}{0.9, 0.98, 1}
\definecolor{highlightgreen}{rgb}{0.9, 0.98, 0.9}
\AtBeginDocument{%
  }

\setcopyright{acmlicensed}
\copyrightyear{2018}
\acmYear{2018}
\acmDOI{XXXXXXX.XXXXXXX}

\acmISBN{978-1-4503-XXXX-X/2018/06}

\renewcommand\footnotetextcopyrightpermission[1]{} 
\acmSubmissionID{9605}
\begin{document}

\title[Leveraging Role-Playing Evaluation in Audio Large Language Models via Reinforcement Learning]{Character Beyond Speech: Leveraging Role-Playing Evaluation in Audio Large Language Models via Reinforcement Learning}

\author{Dongjie Fu}
\authornote{Equal contribution.}
\affiliation{%
  \institution{Zhejiang University}
  \city{Hangzhou}
  \country{China}
}
\email{fudongjie@zju.edu.cn}

\author{Fangming Feng}
\authornotemark[1]
\affiliation{%
  \institution{Zhejiang University}
  \city{Hangzhou}
  \country{China}
}
\email{fangmingfeng@zju.edu.cn}

\author{Xize Cheng}
\authornotemark[1]
\affiliation{%
  \institution{Zhejiang University}
  \city{Hangzhou}
  \country{China}
}
\email{chengxize@zju.edu.cn}

\author{Linjun Li}
\affiliation{%
  \institution{Meituan}
  \city{Shanghai}
  \country{China}
}
\email{lilinjun05@meituan.com}

\author{Zhou Zhao}
\affiliation{%
  \institution{Zhejiang University}
  \city{Hangzhou}
  \country{China}
}
\email{zhaozhou@zju.edu.cn}

\author{Tao Jin}
\authornote{Corresponding author.}
\affiliation{%
  \institution{Zhejiang University}
  \city{Hangzhou}
  \country{China}
}
\email{jint_zju@zju.edu.cn}



\begin{abstract}

The rapid evolution of multimodal large models has revolutionized the simulation of diverse characters in speech dialogue systems, enabling a novel interactive paradigm. Character attributes are manifested not only in textual responses but also through vocal features, as speech conveys rich paralinguistic information that is challenging to quantify. This poses significant difficulties in evaluating the character alignment of role-playing agents. To address these challenges, we present RoleJudge, an evaluation framework that leverages audio large language models to systematically assess the alignment between speech and character across multiple modalities and dimensions. Furthermore, we introduce RoleChat, the first voice role-playing evaluation dataset enriched with chain-of-thought reasoning annotations, comprising a diverse set of authentic and LLM-generated speech samples. Utilizing this dataset, we implement a multi-stage training paradigm and incorporate Standard Alignment in reinforcement learning to mitigate reward misalignment during optimization. Experimental results in terms of accuracy and subjective assessment demonstrate that RoleJudge outperforms various baseline models, validating the effectiveness of our multidimensional evaluation framework.

\end{abstract}




\maketitle

\section{Introduction}

The continuous advancement of artificial intelligence is profoundly transforming the way humans interact with digital systems, giving rise to new forms of digital life that seamlessly integrate technology with human experience. Among these innovations, Role-Playing Agents (RPAs) are particularly noteworthy, as they embody our aspiration to create virtual entities capable of understanding, responding, and interacting with users in increasingly human-like ways. By simulating a wide range of characters, from historical figures and fictional personalities to everyday individuals, these agents open up new possibilities for virtual assistants, interactive storytelling, and intelligent game characters.

Driven by large language models ~\citep{bai2023qwen,dubey2024llama,fang2025llama,yang2025qwen3}, text-based RPAs are becoming a reality ~\citep{shao2023characterllmtrainableagentroleplaying,shanahan2023role,wang2023incharacter}, extending to novel applications such as digital humans and character-driven video games ~\citep{xu2024mindecho}.
With the increasing integration of multimodal technologies and large-scale models ~\citep{speechteam2024funaudiollm,chen2025fireredchat,internlmxcomposer2_5_OL}, a subset of RPAs has begun to prioritize direct human-computer interaction through voice-based communication. ~\citep{zhang2025omnicharacter} Beyond semantic content, spoken language conveys paralinguistic cues, including style and emotion, that are fundamental to expressing the character's personality. Achieving optimal alignment between model-generated outputs and predefined character profiles necessitates producing voice dialogues that faithfully emulate the intended character, thereby enhancing user immersion. Consequently, a critical challenge has emerged: assessing whether the speech generated by RPAs authentically embodies the character and systematically exploring character traits beyond surface-level linguistic content.

\begin{figure*}[t]
\centering
\includegraphics[width=0.96\linewidth]{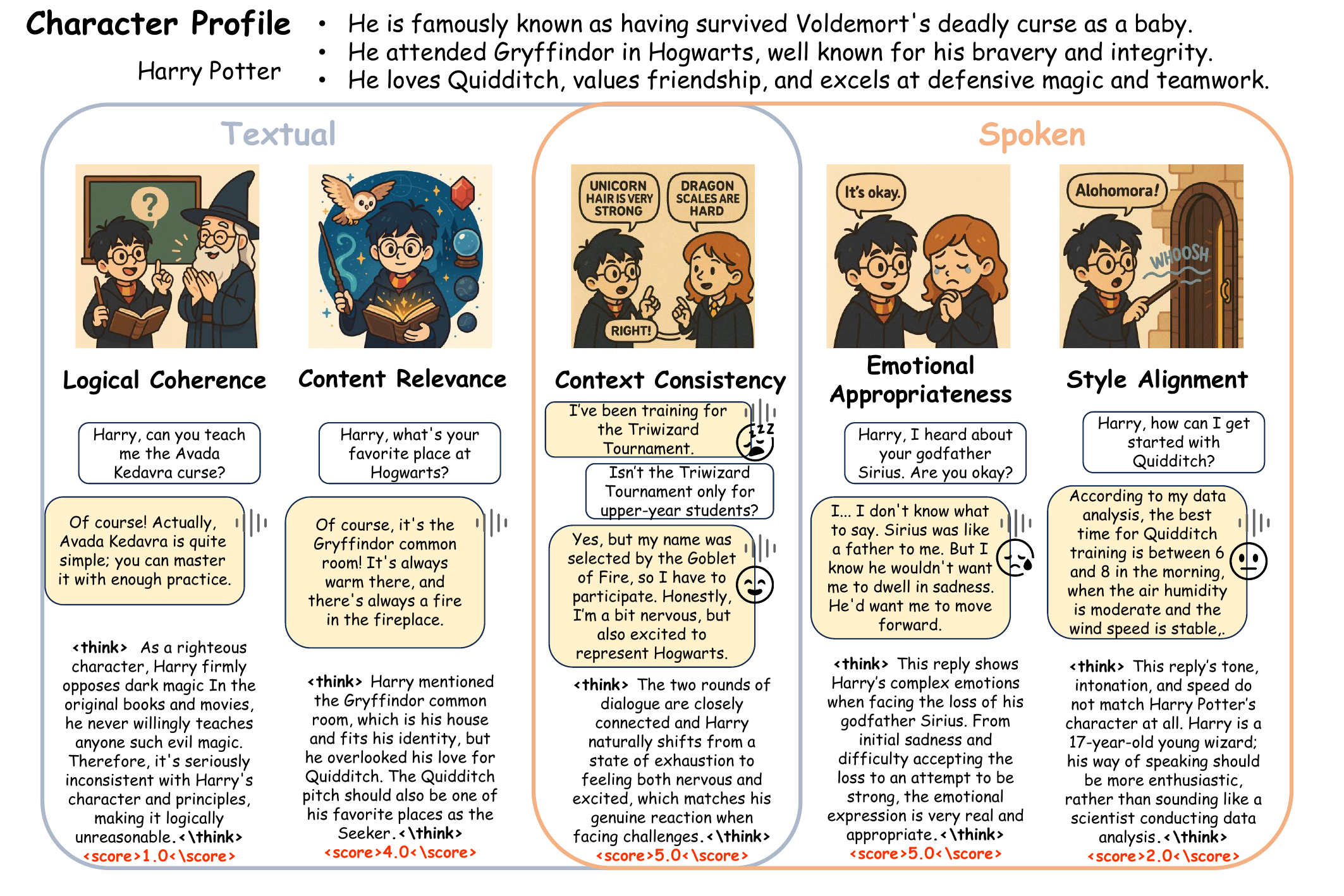}
\caption{RoleChat encompasses five evaluation dimensions: Logical Coherence, which assesses the logical soundness of the response text; Content Relevance, which evaluates whether the response aligns with the character information; Context Consistency, which measures the semantic coherence across multiple dialogue turns as well as the smoothness of emotional transitions; Emotional Appropriateness, which examines the plausibility of the expressed emotions in the response; and Style Alignment, which determines whether the vocal style matches the character.}
\label{fig:fig1}
\end{figure*}

The evaluation of textual outputs generated by RPAs constitutes a vibrant area of research, where authentic character dialogue data sourced from films, novels, and games are utilized to assess agents across dimensions such as interaction capability, character consistency, and user engagement ~\citep{tu2024charactereval,chen2024socialbench,feng2025emocharacter,zhang-etal-2025-roleplot}.
In contrast, spoken language introduces complex acoustic information absent in textual modalities. The nuanced interplay between these acoustic features and character traits renders evaluating voice-based RPAs highly subjective and methodologically challenging. As a result, conventional text-based benchmarks are insufficient for assessing spoken outputs, leaving the evaluation of voice-enabled RPAs an open research problem. Nonetheless, recent advancements in audio foundation models present promising avenues for addressing these challenges.

Audio foundation models are designed for various audio and speech challenges~\citep{tang2023salmonn,xu2025qwen2,chen2024slam,ghosh2025audio,zhang2024internlm}. However, supervised fine-tuning (SFT) on task-specific datasets often constrains their evaluative capabilities, as they are primarily optimized for generation or recognition rather than assessment. Recent efforts have sought to enhance the evaluation capacity of audio models by constructing paired speech-evaluation datasets, targeting applications such as synthetic audio quality assessment~\citep{chen2025audio} and the evaluation of intelligence and emotional quotient in spoken dialogues~\citep{ji2025wavreward}. Despite these advancements, applying such methodologies to RPA evaluation presents two primary challenges: (1) Existing approaches are typically uni-dimensional, yielding a single score that fails to encapsulate the multifaceted nature of speech quality and lacks interpretability; (2) The SFT paradigm inherently limits model generalization, which is essential for handling diverse evaluative tasks.
Furthermore, reinforcement learning-based methods are highly sensitive to data quality. With sparse reward signals, models often deviate from the global optimum and fall into local optima due to insufficient feedback ~\citep{guo2025deepseek}, impairing overall performance.

In the light of these challenges, we introduce RoleChat, the first reasoning-enhanced evaluation dataset for role-playing dialogue, comprising 50 distinct characters and 14,032 samples. The character profiles span a diverse spectrum of personas across various demographics and temperaments, ensuring the dataset's representativeness for real-world scenarios. The dataset consists of both collected and large model-generated samples, with each sample containing character information, dialogue history, user queries, and model outputs. For identical dialogue histories, we sample diverse model outputs to enable a more comprehensive understanding of conversations from multiple perspectives. Each sample is annotated with detailed reasoning and scored across five evaluation dimensions: Logical Coherence, Content Relevance, Context Consistency, Emotional Appropriateness, and Style Alignment, as illustrated in Figure~\ref{fig:fig1}. The quality of both the speech data and evaluation scores is rigorously ensured. Building upon this dataset, we propose a multidimensional evaluation framework, RoleJudge. A subset of RoleChat data is utilized for supervised fine-tuning of audio large models to achieve cold-start initialization, equipping the models with fundamental task comprehension and appropriate output formatting capabilities. Subsequently, we employ standard alignment reinforcement learning, where, based on the GRPO framework ~\citep{guo2025deepseek}, authentic or high-scoring samples are introduced as standards. The model's understanding of these standard samples represents its evaluative performance on corresponding tasks. The average reward of standard samples is used as a scaling parameter for other samples with identical query, preventing the model from selecting relatively high-reward actions in scenarios with low absolute rewards and thus avoiding local optima. Our main contributions are as follows:
\begin{itemize}[left=0em, itemsep=-0.5pt, label=\textbullet]
    \item RoleJudge is the first evaluation model specifically designed for voice-based role-playing dialogue. It takes speech-to-speech conversations as input and assesses the quality of responses from multiple perspectives, including text and speech multimodality, as well as alignment and consistency. Extensive experiments demonstrate the effectiveness of RoleJudge.
    
    \item RoleJudge introduces standard rewards as absolute guidance in positive and negative multi-sample sampling, optimizing the alignment of reward signals under relative reward settings and thereby enhancing the model's evaluative capacity.
    
    \item We present RoleChat, a large-scale, reasoning-enhanced role-playing dialogue evaluation dataset. Alongside diverse synthesized and authentic responses, RoleChat features a purely human-annotated gold-standard evaluation set, ensuring unbiased, high-fidelity assessment of models against genuine human preferences.
    
\end{itemize}



\section{Related Works}

\subsection{Role-Playing Agents.}

Role-Playing Agents (RPAs) are intelligent agents capable of simulating the knowledge, behaviors, emotions, and communication styles of specific characters, thereby achieving highly anthropomorphic role-playing abilities ~\citep{shanahan2023role,shao2023characterllmtrainableagentroleplaying}. RPAs typically leverage capabilities such as in-context learning, instruction following, and social intelligence to reproduce the linguistic and behavioral characteristics of historical figures, fictional characters, or real individuals ~\citep{zhou2024characterglm}. The outstanding performance of large language models (LLMs) in generating human-like content has greatly propelled the development of RPAs. Some works employ retrieval-augmented generation (RAG) and similar methods to enable agents to reproduce character-specific knowledge ~\citep{li2023chatharuhi}, while other studies focus on aligning the linguistic style with the target persona ~\citep{wang2023rolellm}, and yet others aim to train agents with profile and experience perception to reflect deeper personality traits ~\citep{lu2024large}. Recently, with the advancement of multimodal technologies, RPAs have gradually expanded to include multimodal features such as voice style. For example, OmniCharacter seamlessly integrates speech and language to ensure immersive interactions for RPAs ~\cite{zhang2025omnicharacter}.

As the application scope of RPAs continues to expand, the evaluation of LLMs’ role-playing capabilities has garnered significant attention. RoleEval ~\citep{shen2023roleeval} pioneered a bilingual benchmark utilizing multiple-choice queries to gauge character knowledge acquisition, comprehension, and reasoning. Conversely, TimeChara ~\citep{ahn2024timechara} shifts the focus to the agents' capacity for error identification and self-correction. Further advancing evaluative granularity, CharacterEval ~\citep{tu2024charactereval} establishes a multi-dimensional metric framework and introduces CharacterRM, a human-annotated reward model designed to capture subjective nuances in role-playing.

However, these text-centric methodologies are ill-suited for voice-based interaction scenarios, which represent a more direct and prevalent paradigm in practical applications. While VoxRole ~\citep{wu2025voxrolecomprehensivebenchmarkevaluating} has attempted to bridge this gap by assessing the alignment between acoustic features and linguistic style, it exhibits a methodological limitation: it relies on audio models merely for paralinguistic feature extraction, delegating the final automated assessment to text-based models. This approach overlooks the intrinsic evaluative potential of Large Audio Models, which are capable of integrating more granular and effectual acoustic information directly. Consequently, a more comprehensive and native multimodal assessment framework is required for evaluating RPAs.

\subsection{LLMs for Speech Information Perception.}

In recent years, the development of multimodal technologies has enabled the alignment of audio modalities with large model inputs, thereby facilitating extensive audio understanding by large language models. Some studies encode speech into discrete tokens and incorporate them into LLMs, allowing the models to accept audio input, as seen in works such as SpeechGPT ~\citep{zhang2023speechgpt} and AudioPaLM ~\citep{kong2024audio}. Models like SALMONN ~\citep{tang2023salmonn} and Qwen-Audio ~\citep{chu2023qwen,chu2024qwen2} are trained on large-scale, multi-task datasets, equipping them to perform a variety of downstream tasks including speech recognition, speech translation, and audio event detection. A subset of research applies large audio models to spoken dialogue, enabling more intelligent interactions, for example, by mining paralinguistic factors such as style to generate emotionally rich responses ~\citep{lin2024advancing}, or by avoiding cascaded approaches to achieve more real-time interaction. ~\citep{zeng2024glm,vitaaudio}

Recently, studies have explored the potential of large audio models in evaluating speech-related tasks. Specifically, reinforcement learning has been introduced for the first time, utilizing large audio models as descriptive speech quality evaluators to assess TTS outputs and achieve more accurate evaluation ~\citep{chen2025audio}. WavReward ~\citep{ji2025wavreward} further extends this approach by employing chain-of-thought reasoning, using models to evaluate both the intelligence and emotional quotient of end-to-end spoken dialogue systems. These works demonstrate the enhanced generalization capabilities of reinforcement learning in evaluation tasks. However, when facing the multidimensional requirements of role-playing evaluation, the training strategies still require redesign, and high-quality datasets are essential, as annotation errors can undermine the learning of reward signals.

To address these challenges, we have constructed RoleChat, a dataset specifically designed for role-playing dialogue evaluation, encompassing five dimensions of assessment. We introduce reinforcement learning with standard alignment, introducing model performance as an absolute score to scale the advantages within sample groups, thereby reducing the occurrence of selecting the best among suboptimal options.
This approach effectively improves the accuracy of models in role-playing evaluation tasks.


\begin{figure*}
    \centering
    \includegraphics[width=0.96\linewidth]{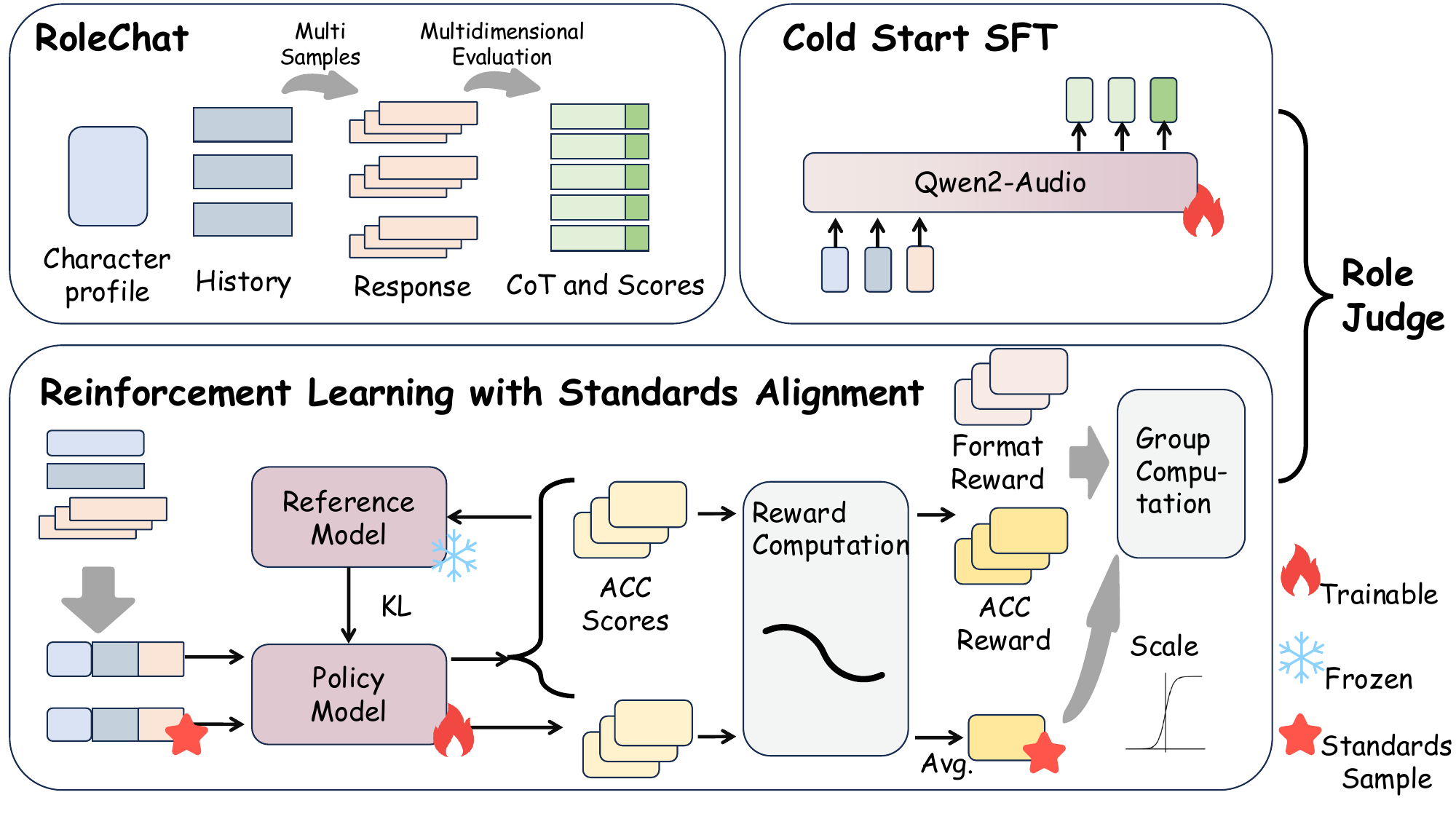}
    \caption{The overall architecture of RoleJudge. It comprises initial model supervised fine-tuning and standard alignment reinforcement learning with multi-dimension. Leveraging an audio large model backbone, RoleJudge facilitates joint understanding of textual and acoustic modalities, thereby enabling fine-grained analysis and holistic assessment of role-playing dialogues.}
    \label{fig:model}
\end{figure*}

\section{RoleJudge: Multidimensional Evaluation Framework}

\subsection{Overview}

Following the training framework of DeepSeek-R1~\citep{guo2025deepseek}, the overall pipeline of Role Judge consists of supervised fine-tuning (SFT) with a subset of data for cold-start initialization, subsequent reinforcement learning-based post-training with standard alignment, as illustrated in Figure \ref{fig:model}. The baseline model for Role Judge is Qwen2-Audio ~\cite{chu2024qwen2}, which demonstrates strong performance across various audio-related tasks. On the input side, the large language model leverages the alignment between the audio encoder and the language model, enabling simultaneous comprehension of both semantic and acoustic information within speech. Compared to cascaded approaches that separately extract audio features and utilize text-based large language models, Qwen2-Audio is better suited for the evaluation of voice-based role-playing agents.

We define the evaluation task for role-playing speech as follows: Given the character profile $P$, the dialogue history sequence $\{h_0,h_1...h_k\}$ between the role and the user, the current user query $q$, and the agent’s response $t$, the evaluation model is required to understand $t$ from both semantic and acoustic perspectives. Integrating all available information, the model must assess the agent’s speech output across five dimensions: response rationality, response consistency, historical coherence, emotional appropriateness, and stylistic alignment. The model should output both the chain-of-thought reasoning process $c_i$ and the final scores $s_i$, with $i$ representing evaluations dimensions. 
For model training, we directly concatenate the encoded representations of textual and audio information as the input, thereby improving the model’s capability for multimodal comprehension.

\subsection{Cold-Start Supervised Fine-Tuning}
To ensure a stable and effective reinforcement learning trajectory, we initiate the training process with a cold-start supervised fine-tuning (SFT) phase. During this stage, the model is optimized to minimize the negative log-likelihood of generating target outputs, utilizing a curated dataset of paired audio-text samples rich in chain-of-thought reasoning and multidimensional quality metrics. This phase is instrumental in equipping the model with a foundational grasp of complex evaluation logic and the requisite structured formatting, thereby establishing a high-fidelity starting policy that facilitates more efficient exploration and robust optimization during the subsequent reinforcement learning.

\subsection{Reinforcement Learning with Standard Alignment}

In large-scale model training, reinforcement learning (RL) methods are widely utilized to align model outputs with human preferences and optimize generation quality. Classic algorithms such as Proximal Policy Optimization (PPO) \citep{schulman2017proximal,yu2022surprising}, which relies on a separately trained value function (critic), and Direct Preference Optimization (DPO) \citep{rafailov2023direct}, which leverages contrastive preference signals, have achieved remarkable success in text domains. However, applying these approaches directly to multimodal role-playing speech evaluation poses significant challenges. Specifically, the nuanced and multidimensional nature of speech (encompassing tone, emotion, and paralinguistic cues) makes it exceedingly difficult to define a stable scalar value function or consistently align discrete preference signals.

To address these challenges, we adopt Group Relative Policy Optimization (GRPO) \citep{guo2025deepseek} as our base RL framework. GRPO introduces a group-based sampling paradigm that inherently models the relative variance among multiple outputs, bypassing the need for an external critic model. For a given evaluation query, we sample a group of $G$ candidate responses. The mean reward within this group serves as the dynamic baseline, and the relative advantage of each sample is used to update the policy. 

Considering that RoleJudge is required to generate both the chain-of-thought reasoning process $c$ and the final multidimensional scores $s$, we define the base reward function as an aggregation of two critical components: the format reward $r_f$ and the accuracy reward $r_a$. The format reward $r_f \in \{0, 1\}$ acts as a hard constraint, strictly enforcing adherence to the required structural tags. The accuracy reward $r_a$, inspired by recent advancements in speech evaluation metrics \citep{ji2025wavreward}, utilizes a Gaussian-like non-linear decay to penalize deviations from the human-annotated score $s_c$:
\begin{equation}
r_a(s, s_c) = 10 \cdot \exp\left( -\frac{(s_c - s)^2}{2\sigma^2} \right)
\end{equation}
where $\sigma$ controls the tolerance width of the scoring discrepancy. This exponential formulation smoothly encourages the policy to converge toward exact accuracy.

A critical vulnerability of standard GRPO in complex reasoning tasks is its susceptibility to ``choosing the best among the worst.'' When the policy model fails to comprehend the task and generates universally poor outputs for a specific query, normalizing the rewards within this low-quality group still assigns positive advantages to slightly less erroneous outputs. This spurious signal can easily trap the policy in local optima. 

To mitigate this reward misalignment, we introduce a novel \textbf{Standard Alignment} mechanism. A unique advantage of role-playing datasets like RoleChat is the availability of authentic, high-quality reference data (standard samples) mined directly from real-world scenarios. We hypothesize that a model's ability to accurately evaluate these ground-truth standard samples serves as an absolute indicator of its current comprehension level for the given query. 

Specifically, during each RL iteration, before evaluating the $G$ generated candidates, the policy model is first prompted to evaluate $M$ standard samples associated with the same query. We calculate the average accuracy reward on these standard samples, denoted as $r_u$. This $r_u$ acts as a confidence proxy. We subsequently utilize $r_u$ to dynamically scale the advantage estimation for the $G$ candidate samples:
\begin{equation}
A_i = \phi(r_u) \frac{r_i - \mu_r}{\sigma_r + \epsilon_{std}} \quad \text{for } i \in \{1, \dots, G\}
\end{equation}
where $\mu_r$ and $\sigma_r$ are the empirical mean and standard deviation of the candidate rewards $r_1, \dots, r_G$, and $\epsilon_{std}$ is a small constant for numerical stability. The scaling factor $\phi(r_u)$ is defined as a smooth sigmoid transition:
\begin{equation}
\phi(r_u) = a + (b-a) \cdot \text{sigmoid}(\alpha(r_u - 0.5))
\end{equation}
Here, $a$ and $b$ govern the lower and upper bounds of the scaling factor, and $\alpha$ dictates the sharpness of the transition. In essence, if the model scores poorly on the standard samples (low $r_u$), $\phi(r_u)$ shrinks the advantage $A_i$. This conservatively reduces the magnitude of the policy update, preventing the model from blindly optimizing based on noisy relative rankings.

Furthermore, we employ $r_u$ to dynamically re-weight the optimization focus between structural correctness and scoring precision for the total candidate reward $R_i$:
\begin{equation}
R_i = \lambda(r_u) \cdot r_{a,i} + (1 - \lambda(r_u)) \cdot r_{f,i}
\end{equation}
where $\lambda(r_u)$ is a monotonically increasing function of $r_u$. Intuitively, when task comprehension is poor (low $r_u$), the objective shifts toward $r_f$, ensuring the model at least learns to maintain formatting stability. As comprehension improves (high $r_u$), $\lambda(r_u)$ increases, prompting the model to focus rigorously on refining its evaluation accuracy.

Following the core GRPO architecture, the final objective integrates the scaled advantage $A_i$ and a Kullback-Leibler (KL) divergence penalty to ensure training stability. The loss function to be minimized is formulated as:
\begin{equation}
\begin{split}
\mathcal{L} = - \frac{1}{G} \sum_{i=1}^G \bigg[ & \min \left( \rho_i A_i, \text{clip}(\rho_i, 1-\epsilon, 1+\epsilon) A_i \right) \\
& - \beta D_{KL}(\pi_\theta(o_i) \| \pi_{\text{ref}}(o_i)) \bigg]
\end{split}
\end{equation}
where $\rho_i = \pi_\theta(o_i)/\pi_{\text{old}}(o_i)$ denotes the probability ratio, $\epsilon$ restricts excessively large policy updates, $\pi_{\text{ref}}$ is the frozen reference model, and $\beta$ is the regularization coefficient. The complete training procedure is summarized in Algorithm \ref{alg:standard_alignment}.

\begin{algorithm}[htbp]
\caption{RL with Standard Alignment for RoleJudge}
\label{alg:standard_alignment}
\begin{algorithmic}[1]
\REQUIRE Policy $\pi_\theta$, Reference $\pi_{\text{ref}}$, Dataset $\mathcal{D}$, Hyperparameters $G, M, a, b, \alpha$.
\FOR{each training iteration}
    \STATE Sample query $q \sim \mathcal{D}$, fetch $M$ standard samples $x_{std}$, and generate $G$ candidates $x_{cand}$.
    \STATE Evaluate $o_{std} \sim \pi_\theta(\cdot | q, x_{std})$ to get standard accuracy rewards $r_{a, std}$ (Eq. 1).
    \STATE Compute confidence $r_u = \frac{1}{M} \sum_{m=1}^M r_{a, std}^{(m)}$; derive $\phi(r_u)$ (Eq. 3) and $\lambda(r_u)$.
    \STATE Evaluate $o_{cand} \sim \pi_\theta(\cdot | q, x_{cand})$ to get candidate rewards $r_{f,i}$ and $r_{a,i}$.
    \STATE Aggregate rewards $R_i = \lambda(r_u) r_{a,i} + (1 - \lambda(r_u)) r_{f,i}$ for all $i \in \{1 \dots G\}$.
    \STATE Estimate scaled advantages $A_i = \phi(r_u) \frac{R_i - \mu_R}{\sigma_R + \epsilon_{std}}$ using group mean $\mu_R$ and std $\sigma_R$.
    \STATE Update $\theta$ by minimizing the GRPO loss $\mathcal{L}$ (Eq. 5).
\ENDFOR
\end{algorithmic}
\end{algorithm}
\begin{table*}[t]
\caption{Accuracy performance of RoleJudge and other baselines on RoleChat across multi evaluation dimensions: Logical Coherence (L-C), Content Relevance (C-R), Context Consistency (C-C), Emotional Appropriateness (E-A), and Style Alignment (S-A), Overall Acc and Format Acc. The bolded scores indicate the best performance achieved in each respective dimension.}
\label{tab:main_results}

\centering 
\begin{tabularx}{0.95\textwidth}{@{} ll *{7}{>{\centering\arraybackslash}X} @{}}
\toprule

\multicolumn{2}{c}{\multirow{2}{*}{\textbf{Method}}} & \multicolumn{2}{c}{\textit{\textbf{Textual}}} 
& \multicolumn{2}{c}{\textit{\textbf{Spoken}}}
& \multicolumn{1}{c}{\multirow{2}{*}{\textbf{C-C}}}
& \multicolumn{1}{c}{\multirow{2}{*}{\textbf{Overall Acc}}}
& \multicolumn{1}{c}{\multirow{2}{*}{\textbf{Format Acc}}}
\\ 
\cmidrule(l){3-4} \cmidrule(l){5-6} 

&& \multicolumn{1}{c}{\textbf{L-C}}                
& \multicolumn{1}{c}{\textbf{C-R}}                  
& \multicolumn{1}{c}{\textbf{E-A}}              
& \multicolumn{1}{c}{\textbf{S-A}}                  
&  & &\\ \midrule
\multicolumn{1}{c}{\multirow{5}{*}{\textbf{\shortstack{\textbf{Text-}\\\textbf{Modality}}}}}&\multicolumn{8}{l}{\cellcolor[HTML]{EFEFEF}\textit{Open-Source Models}} \\
& Qwen3-8B   & 62.5 & 42.1 & 18.5 & 12.3 & 32.1 & 33.50 & 91.1 \\
& Qwen3-32B  & 66.2 & 46.5 & 21.2 & 14.6 & 35.4 & 36.78 & 94.3 \\
&\multicolumn{8}{l}{\cellcolor[HTML]{EFEFEF}\textit{Closed-Source Models}} \\
& GPT-4.1      & \textbf{96.6} & \textbf{92.1} & 28.4 & 19.5 & 48.2 & 56.96 & \textbf{100} \\
\midrule
\multicolumn{1}{c}{\multirow{9}{*}{\textbf{\shortstack{\textbf{Multi-}\\\textbf{Modality}}}}}&\multicolumn{8}{l}{\cellcolor[HTML]{EFEFEF}\textit{Open-Source Models}} \\
& SALMONN-7B   & 11.2 & 23.2 & 43.2 & 12.1 & 22.1 & 22.36 & 6.2 \\
& Qwen-Audio   & 35.2 & 29.3 & 34.2 & 16.2 & 32.3 & 29.48 & 0 \\
& Qwen2-Audio  & 40.9 & 25.1 & 42.1 & 11.1 & 34.1 & 30.66 & 10.2 \\
& Qwen3-Omni   & 63.8 & 43.3 & 51.6 & 22.1 & 35.5 & 43.26 & 75.8 \\
&\multicolumn{8}{l}{\cellcolor[HTML]{EFEFEF}\textit{Closed-Source Models}} \\
& GPT-4o-audio & 65.2 & 42.3 & 61.2 & 52.4 & 44.2 & 53.06 & 94.2 \\
& Gemini3 Pro  & 86.5 & 72.9 & 75.8 & 51.6 & 62.2 & 69.80 & \textbf{100} \\
\cmidrule(l){2-9}
& RoleJudge    & 94.8 & 90.2 &  \textbf{85.1} &  \textbf{75.9} &  \textbf{84.0} &  \textbf{86.00} & \textbf{100} \\
\bottomrule
\end{tabularx}
\end{table*}

\section{RoleChat: Constructing a Reasoning-Rich Dataset for Dialogue Evaluation}
\subsection{Overall}
To enable models to accurately assess the quality of role-playing speech from multiple dimensions, we present role-chat, a evaluation dataset encompassing role-playing dialogues. This dataset features comprehensive character profiles and provides diverse responses—including both positive and negative examples—for identical scenarios, as well as a subset of real speech data. 
Each dialogue sample is annotated with multi-dimensional reasoning and scoring. Crucially, while the training corpus utilizes a scalable hybrid annotation pipeline, we deliberately construct a purely human-annotated gold-standard evaluation set. To ensure the high quality of the entire dataset, we have established a rigorous and systematic data construction pipeline.

\subsection{Dataset Construction}
\textbf{Stage 1: Character Profile Construction.}
To collect authentic speech data, we curate 50 virtual characters from films, television dramas, and other audiovisual works. To ensure the uniqueness of each character profile, we conducted a detailed summary of their personal information. We gather background information, key plot points, and selected lines from these works, and leverage the powerful generative capabilities of large language models ~\citep{openai2024gpt4technicalreport} to extract and summarize character details, forming comprehensive profiles that include personality traits, experiences, hobbies, and habits. Subsequently, all profiles are manually verified and any unfaithful information was removed to ensure the accuracy of character identities.

\textbf{Stage 2: Dialogue Text Generation.}
For the generation of textual dialogues, we adopted a dual approach to construct dialogue histories and user queries. One approach involves collecting authentic dialogue histories directly from film and television works, ensuring the data reflects real-world scenarios and remains faithful to the character’s persona. The other approach utilizes synthetic historical scenarios, where we employ GPT-4.1 ~\citep{openai2024gpt4technicalreport} to generate plausible interactions between characters and users, covering a wide range of topics such as daily life, character experiences, and personal viewpoints. We explicitly require that character utterances do not contradict their profiles, thereby guaranteeing the accuracy of the dialogue history. For the final character responses, the segments to be evaluated, we use models from the Qwen2.5 series ~\citep{bai2023qwen} of various sizes, as well as the GPT series ~\citep{gpt4o,openai2024gpt4technicalreport}, to generate diverse replies, sampling a range of response qualities to enrich the evaluation dataset.

\textbf{Stage 3: Dialogue Speech Generation.}
During the speech dialogue generation phase, for synthetic historical scenarios, we leverage existing character audio and apply zero-shot TTS with CosyVoice ~\citep{du2024cosyvoice} to construct character speech for the dialogue history. For character responses, we randomly select different audio samples from the same character, from other characters, or use the TTS model’s default voice settings with randomly assigned emotions, intonation, speed, and accent to generate a variety of speech samples, thereby maximizing acoustic diversity. Since the reference audio already contains attributes such as emotion and character style, randomly selecting reference samples enables the construction of speech outputs with diverse styles. Additionally, incorporating audio from different characters and instruction-based TTS further enriches the stylistic diversity of the samples. After generating speech samples, we employ the SenseVoice model ~\citep{speechteam2024funaudiollm} for ASR and filter out samples with high WER to ensure the quality of the synthesized speech.

\textbf{Stage 4: Data Scoring and Standard Selection.} 
For sample reasoning and scoring, we employ a cascaded annotation pipeline that deliberately decouples audio perception and logical deduction. Specifically, we leverage Gemini-3 Pro ~\citep{comanici2025gemini} to extract fine-grained acoustic descriptions, such as emotion and prosody. Based on these multimodal features, GPT-4.1\citep{openai2024gpt4technicalreport} generates rigorous reasoning chains and final scores. 

To ensure the reliability of these machine-generated annotations, trained volunteers conducted a Human-in-the-Loop (HITL) verification. During this process, we systematically identified ``standard samples'' to serve as absolute reference anchors for our RL alignment. For authentic dialogues, the original real-world responses were directly designated as standard anchors. For synthetic scenarios, annotators reviewed generated samples that achieved perfect scores across all dimensions, manually selecting the single response that most authentically embodied the character's persona. 

Finally, to rigorously evaluate RoleJudge, we partitioned 10\% of the dataset as the evaluation set. Crucially, this subset was entirely annotated from scratch by human evaluators, bypassing the automated pipeline. This strategic separation ensures that our evaluation reflects genuine human perception and mitigates the risk of the model merely overfitting to the idiosyncratic scoring preferences of the teacher models.

\section{Experiments}
\subsection{Datasets and Baselines.}
Regarding the dataset split, the aforementioned 10\% human-annotated evaluation set is carefully curated to include three roles completely unseen during training, alongside a standard 3\% validation set. This deliberate partitioning allows us to rigorously assess the model's generalization capabilities to novel characters. Furthermore, the evaluation data for every role explicitly incorporates dialogues drawn directly from real-world scenarios, enabling us to verify the model's evaluative accuracy and robustness in authentic, non-synthetic settings.

To comprehensively evaluate the role-playing assessment capability of RoleJudge, we compared multiple large model-based approaches across different modalities, model sizes, and architectures. These include single-text modality open-source models such as Qwen3-8B and Qwen3-32B ~\citep{yang2025qwen3}, as well as the proprietary GPT-4.1 ~\citep{openai2024gpt4technicalreport}, which are used to specifically assess role-playing evaluation from a text perspective (with SenseVoice ~\citep{speechteam2024funaudiollm} ASR results as input). For multimodal audio models, we included open-source models such as SALMONN, Qwen-Audio ~\citep{chu2023qwen}, Qwen2-Audio ~\citep{chu2024qwen2}, and Qwen3-Omni ~\citep{xu2025qwen2}, as well as proprietary models GPT-4o-Audio ~\citep{gpt4o} and Gemini3 Pro ~\citep{comanici2025gemini}.

For evaluation metrics, we primarily adopted accuracy to measure the discrepancy between the predicted scores and the annotated scores. We assessed the model's understanding ability from five dimensions: Logical Coherence, Content Relevance, Context Consistency, Emotional Appropriateness, and Style Alignment. To complement accuracy and provide a more fine-grained assessment, we additionally incorporated Mean Squared Error (MSE) to quantify the absolute magnitude of scoring errors, and the Pearson Correlation Coefficient ($r$) to measure the trend alignment between model predictions and human annotations, particularly for subjective dimensions like Emotional Appropriateness and Style Alignment. We also calculated the average accuracy to evaluate the model's overall capability, and a format accuracy metric to assess whether the model can follow instructions and generate the correct reasoning and evaluation structure. Furthermore, we invited volunteers to participate in our data construction process, generating dialogue data through real-time interactions and conducting A/B testing of the evaluation models.

\subsection{Experimental Setup}
We implemented the RolePlaying multidimensional evaluation framework based on the Qwen2-Audio-7B-Instruct model. The training process is divided into two stages: \textbf{Cold-Start Phase}: This phase aims to enable the model to understand the task and generate reasoning and scores in the correct format. The learning rate is set to $1 \times 10^{-5}$, the batch size is 4, and training is performed on 8 A100 GPUs. \textbf{Reinforcement Learning Phase}: In this phase, we expect the model to accurately comprehend and evaluate speech data across different dimensions. We train five expert models independently, with hyperparameters set as a learning rate of $5 \times 10^{-7}$, batch size of 2, scaling hyperparameters $a=0.5$, $b=1.5$, $\alpha=8$, and $\lambda=0.8$, as well as a KL-divergence regularization beta value of $0.01$. Training is performed on $32$ A100 GPUs.

\begin{figure*}[t]
    \centering
    \begin{subfigure}[b]{0.48\linewidth}
        \centering
        \includegraphics[width=\linewidth]{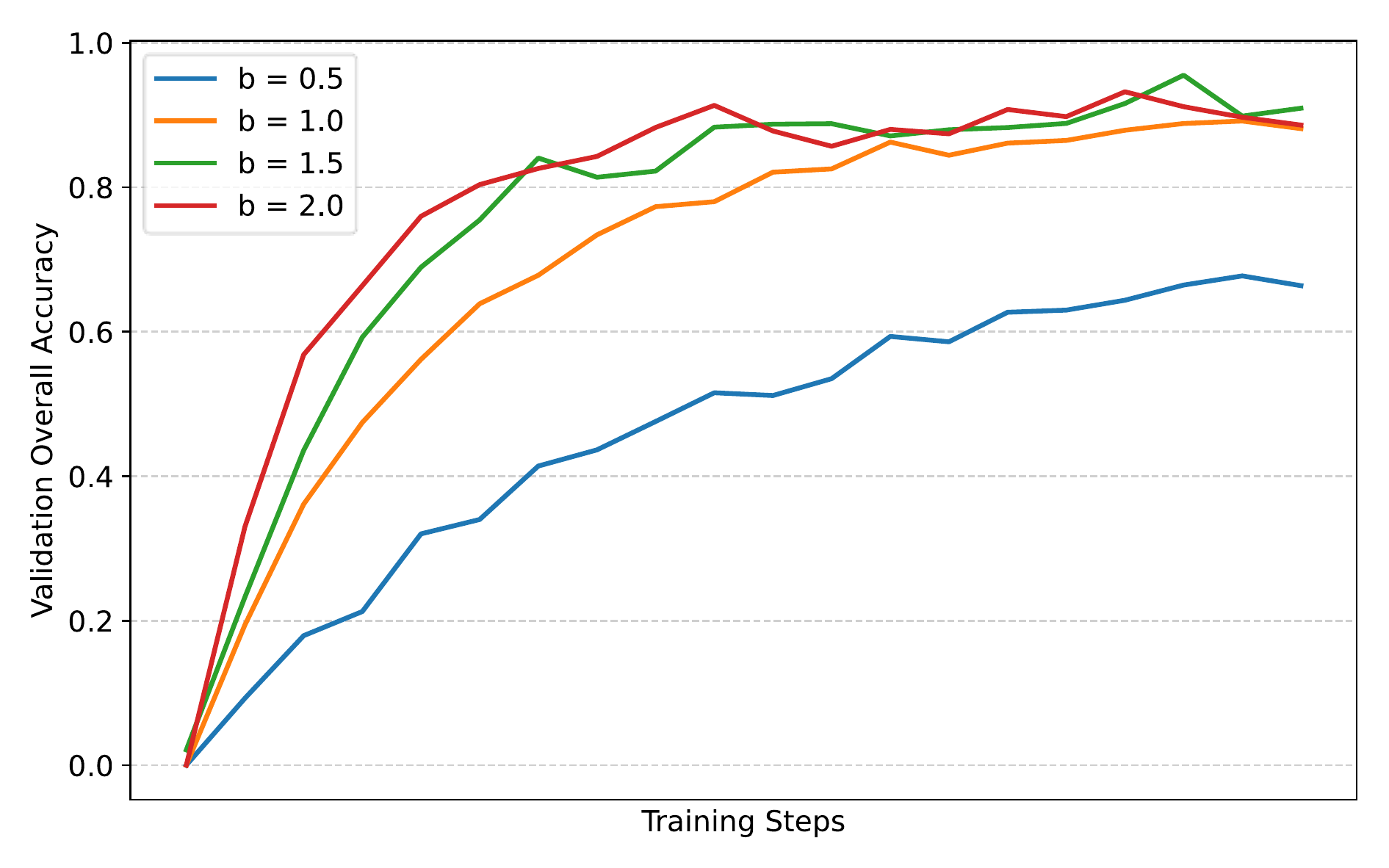}
        \caption{Impact of scaling factor $b$}
        \label{fig:hyper:b}
    \end{subfigure}
    \hfill
    \begin{subfigure}[b]{0.46\linewidth}
        \centering
        \includegraphics[width=\linewidth]{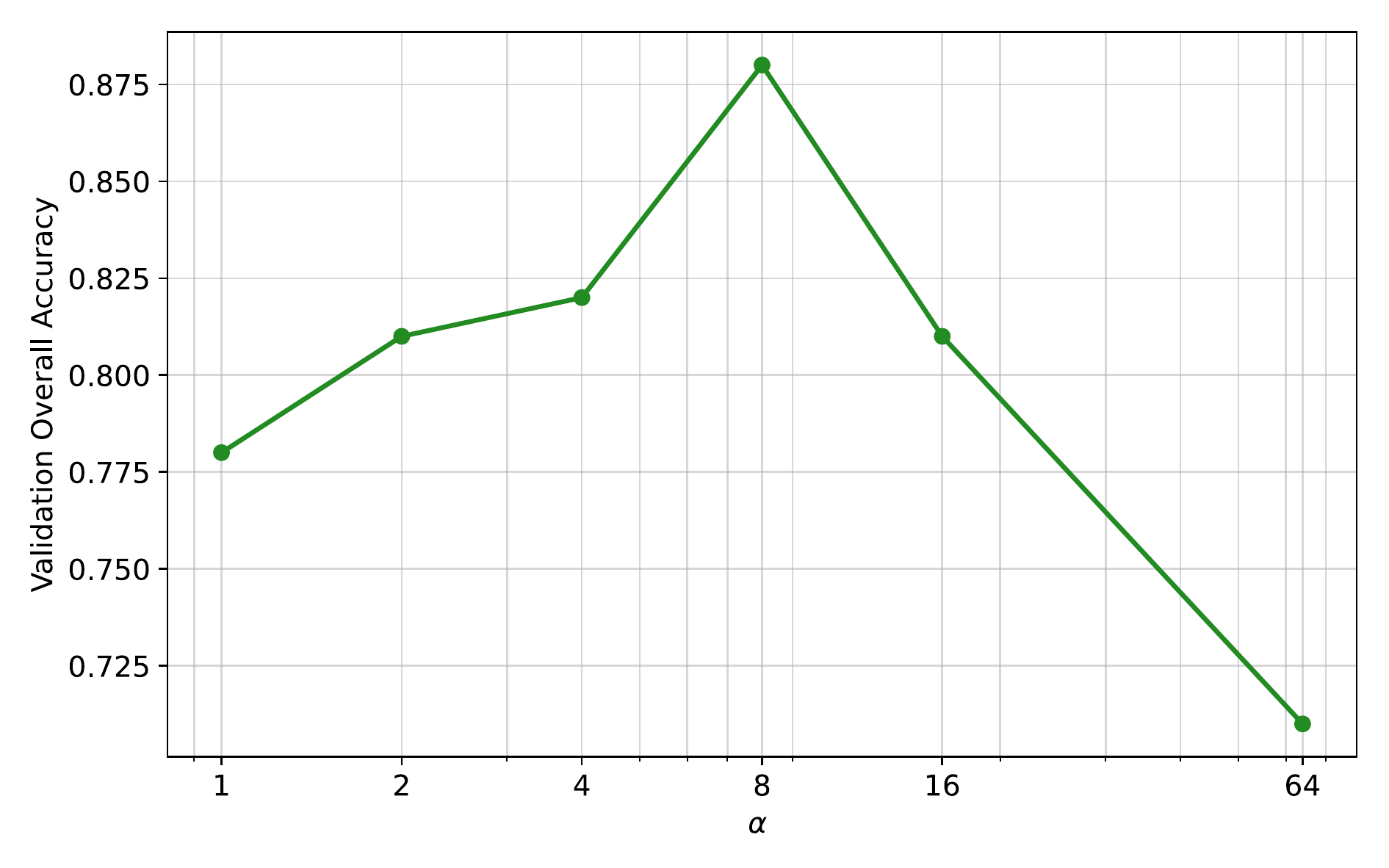}
        \caption{Sensitivity of sharpness $\alpha$}
        \label{fig:hyper:alpha}
    \end{subfigure}
    \caption{Hyperparameter sensitivity analysis of the Standard Alignment mechanism on the validation set.}
    \label{fig:HyperparameterAnalysis}
\end{figure*}
\subsection{Main Results}


As illustrated in Table \ref{tab:main_results}, RoleJudge achieves the best overall evaluation results, surpassing all baseline models across different modalities. A key observation is the severe performance collapse of text-only models when transitioning from linguistic to paralinguistic tasks. Although text-modality models like GPT-4.1 demonstrate superior performance in semantic-heavy dimensions such as Logical Coherence (96.6\%) and Content Relevance (92.1\%), their accuracy drops drastically to near-random guessing levels in spoken-related dimensions. For instance, in Style Alignment (S-A) and Emotional Appropriateness (E-A), GPT-4.1 only achieves 19.5\% and 28.4\% respectively, even with high-quality ASR transcriptions. This catastrophic failure stems from the inherent "acoustic blindness" of text models, which cannot perceive critical paralinguistic cues such as timbre, intonation, and emotional prosody. This disparity strongly underscores that role-playing evaluation is a holistic multimodal task where acoustic fidelity is as critical as linguistic logic, thus fully justifying the necessity of our end-to-end audio evaluation framework.

Beyond exact match accuracy, Table \ref{tab:mse_pearson} provides a more granular assessment of the models' reliability and alignment with human perception. RoleJudge achieves the lowest Overall MSE (0.21), indicating that its scoring deviations are marginal and far more stable than those of proprietary baselines like GPT-4o-audio (1.42). More importantly, because our evaluation set is strictly human-annotated from scratch, the metrics reflect genuine human aesthetics. In the highly subjective dimensions of E-A and S-A, RoleJudge demonstrates strong positive correlations with these pure human annotations, with Pearson coefficients ($r$) reaching 0.81 and 0.62, respectively. This significantly outperforms the strongest baseline, Gemini3 Pro ($r=0.68$ and $0.59$), proving that our Standard Alignment RL mechanism effectively enables the model to capture the nuanced trends of human aesthetic judgment rather than merely outputting discrete values.

\begin{table}[h]
  \centering
  \caption{Evaluation of error magnitude (MSE) and human-alignment correlation (Pearson's $r$) on subjective dimensions.}
  \label{tab:mse_pearson}
  \resizebox{\columnwidth}{!}{
    \begin{tabular}{lccc}
    \toprule
    \multirow{2}{*}{\textbf{Method}} & \textbf{Error $\downarrow$} & \multicolumn{2}{c}{\textbf{Pearson ($r$) $\uparrow$}} \\
    \cmidrule(lr){2-2} \cmidrule(lr){3-4}
    & \textbf{Overall MSE} & \textbf{E-A} & \textbf{S-A} \\
    \midrule
    Qwen2-Audio & 2.94 & 0.35 & 0.26 \\
    Qwen3-Omni & 1.86 & 0.44 & 0.38 \\
    GPT-4o-audio & 1.42 & 0.51 & 0.46 \\
    Gemini3 Pro & 0.68 & 0.68 & 0.59 \\
    \textbf{RoleJudge} & \textbf{0.21} & \textbf{0.81} & \textbf{0.62} \\
    \bottomrule
    \end{tabular}
  }
\end{table}
\vspace{-0.3cm}

\subsection{A/B Test for RoleJudge}
A/B testing is a common subjective evaluation method in which human listeners compare two output results and select the one with higher quality. We recruited ten volunteers who, following a process similar to our data construction, interacted with randomly selected models and randomly assigned TTS role-playing agents to generate ten samples each. These samples were then evaluated and scored by RoleJudge, Qwen3-Omni, and Gemini3 Pro. The volunteers performed pairwise comparisons based on the evaluation results and selected the higher-quality option. As shown in Table \ref{tab:ab}, RoleJudge achieved a significant advantage over the other two models, indicating that its scoring system demonstrates superior performance in real-world scenarios.

\begin{table}[hb]
\centering 
\caption{A/B Test result for RoleJudge.}
\label{tab:ab}
\begin{tabularx}{0.9\linewidth}{>{\centering\arraybackslash}X >{\centering\arraybackslash}X >{\centering\arraybackslash}X} 
\toprule
Models & RoleJudge Win $\uparrow$ & Lose $\downarrow$\\
\midrule
Qwen3-Omni& 87 & 13\\
Gemini3 Pro& 79 & 21\\
\bottomrule
\end{tabularx}
\end{table}

\subsection{Ablation Study}

To evaluate the individual contribution of each core component in our training framework, we conduct an ablation study across three configurations: (1) the full RoleJudge model (SFT + RL with Standard Alignment), (2) GRPO training without the Standard Alignment mechanism, and (3) a baseline using only Supervised Fine-Tuning (SFT).

As summarized in Table \ref{tab:abla}, each stage of our training paradigm is critical for achieving high-fidelity role-playing evaluation. The transition from a purely supervised model to a reinforcement learning framework yields a significant 13.62-point improvement in Overall Accuracy. This substantial gain demonstrates the effectiveness of RL in enhancing the model's generalization capabilities across diverse role-playing scenarios. Furthermore, the integration of Standard Alignment provides an additional performance boost of approximately 3.29 points, confirming its ability to mitigate reward misalignment by providing stable behavioral anchors. Notably, while the SFT baseline occasionally struggles with structural constraints (85.2\% Format ACC), both RL-based variants achieve a perfect 100\% Format Accuracy, indicating that the reinforcement learning process significantly reinforces the model's adherence to complex output instructions.

\begin{table}[h]
\centering 
\caption{Ablation experiments for RoleJudge. R-L denotes Reinforcement Learning and S-A denotes Standard Alignment.}
\label{tab:abla}
\begin{tabularx}{0.8\linewidth}{c c >{\centering\arraybackslash}X >{\centering\arraybackslash}X}
\toprule
R-L & S-A & Overall ACC & Format ACC \\
\midrule
\ding{52} & \ding{52} & \textbf{86.00} & \textbf{100.0} \\
\ding{52} & \ding{56} & 82.71 & 100.0 \\
\ding{56} & \ding{56} & 69.09 & 85.2 \\
\bottomrule
\end{tabularx}
\vspace{-0.3cm}
\end{table}

\subsection{Hyperparameter Sensitivity Analysis}

To further investigate the robustness of the proposed Standard Alignment mechanism, we conduct a sensitivity analysis on two core hyperparameters: the maximum scaling factor $b$ and the sharpness parameter $\alpha$. All results in this section are reported based on the validation set of RoleChat.

As illustrated in Figure \ref{fig:HyperparameterAnalysis}(a), the parameter $b$ significantly influences the optimization efficiency. A higher $b$ amplifies the advantage signals for samples that align well with standard anchors, thereby accelerating the initial convergence. However, we found that $b=1.5$ strikes the best balance between training acceleration and long-term stability, preventing potential oscillations in the later stages of reinforcement learning. 

Regarding the sharpness parameter $\alpha$, Figure \ref{fig:HyperparameterAnalysis}(b) reveals a non-monotonic trend in performance. The validation accuracy peaks at $\alpha=8$, suggesting that a moderate sigmoid transition is optimal for distinguishing task difficulty. An excessively sharp transition (i.e., high $\alpha$) leads to a near-step function that makes the advantage estimation overly sensitive to minor fluctuations in standard rewards, ultimately resulting in unstable gradients and a slight degradation in final accuracy.

\vspace{-0.2cm}
\section{Conclusion}
In this paper, we presented RoleChat, the first multimodal role-playing evaluation dataset enriched with multi-dimensional reasoning annotations. To effectively harness this resource, we developed a robust multi-stage training paradigm for RoleJudge, transitioning from cold-start supervised fine-tuning to reinforcement learning. Crucially, we introduced a novel Standard Alignment mechanism within the RL framework, which dynamically scales advantage estimates to mitigate reward misalignment and ensure optimization stability. Comprehensive empirical evaluation and human A/B testing validate the superiority of our approach over existing baselines. Ultimately, this work provides a foundational benchmark and methodology, paving the way for the development of more authentic and immersive voice-based role-playing agents.

\bibliographystyle{ACM-Reference-Format}
\bibliography{sample-base}

\end{document}